\documentclass[preprint,12pt, a4paper]{elsarticle}

\usepackage{amssymb}
\usepackage{hyperref}
\usepackage{amsmath}
\usepackage{amsfonts}

\usepackage{booktabs}

\usepackage{xcolor}

\journal{Software Impacts}

\begin{document}

\begin{frontmatter}

\title{\emph{BNNpriors}: A library for Bayesian neural network inference with different prior distributions}

\author[eth]{Vincent Fortuin\fnref{eq1}\corref{cor1}}
\author[cam]{Adri\`a Garriga-Alonso\fnref{eq1}}
\author[imp]{Mark van der Wilk\fnref{eq2}}
\author[bris]{Laurence Aitchison\fnref{eq2}}

\address[eth]{ETH Z\"urich, Z\"urich, Switzerland}
\address[cam]{University of Cambridge, Cambridge, UK}
\address[imp]{Imperial College London, London, UK}
\address[bris]{University of Bristol, Bristol, UK}

\fntext[eq1]{Equal contribution.}
\fntext[eq2]{Equal contribution.}

\cortext[cor1]{Corresponding author. Email: \texttt{fortuin@inf.ethz.ch}}

\begin{abstract}

Bayesian neural networks have shown great promise in many applications where calibrated uncertainty estimates are crucial and can often also lead to a higher predictive performance. However, it remains challenging to choose a good prior distribution over their weights. While isotropic Gaussian priors are often chosen in practice due to their simplicity, they do not reflect our true prior beliefs well and can lead to suboptimal performance. Our new library, \emph{BNNpriors}, enables state-of-the-art Markov Chain Monte Carlo inference on Bayesian neural networks with a wide range of predefined priors, including heavy-tailed ones, hierarchical ones, and mixture priors. Moreover, it follows a modular approach that eases the design and implementation of new custom priors. It has facilitated foundational discoveries on the nature of the cold posterior effect in Bayesian neural networks and will hopefully catalyze future research as well as practical applications in this area.

\end{abstract}


\end{frontmatter}

\noindent

\section{Introduction}
\label{sec:intro}

While standard neural networks fit a single point estimate of their weights $\hat{\theta}$ (typically using stochastic gradient descent), Bayesian neural networks (BNNs) instead infer a \emph{posterior distribution} $p(\theta | \mathcal{D})$ over their weights given some training data $\mathcal{D}$ \cite{mackay1992practical, neal1995bayesian}.
They perform this inference using Bayes' theorem, that is, $p(\theta | \mathcal{D}) = Z^{-1} \, p(\mathcal{D} | \theta) \, p(\theta)$ \cite{bayes1763essay}.
Using this posterior, BNNs can then offer a predictive distribution over outputs, which can help quantify their uncertainty with respect to specific predictions \cite{ovadia2019can, ciosek2019conservative}. In safety-critical applications like medicine or autonomous driving, such calibrated uncertainty estimates can be crucial \cite{kendall2017uncertainties}.

The distribution $p(\theta)$ in the equation above is called the \emph{prior distribution} and has to be carefully chosen to encode our true prior beliefs in order to build a successful Bayesian model \cite{gelman2013bayesian}.
However, this is challenging in BNNs, since we often do not know how to properly formulate prior beliefs over the weight space.
In practice, isotropic Gaussian priors are thus often used for their simplicity and computational appeal \cite[e.g.,][]{wilson2020bayesian, dusenberry2020efficient}.

Recently, it has been hypothesized that these isotropic Gaussian priors are the culprit for the cold posterior effect, that is, the fact that tempered posteriors at lower temperatures can perform better than the true Bayesian posterior \cite{wenzel2020good}.
This hypothesis has been partially confirmed and it has been shown that indeed the performance of BNNs can be improved and the cold posterior effect alleviated when using different priors \cite{fortuin2021bayesian}.
Specifically, heavy-tailed priors seem to work well for fully-connected BNNs and correlated priors for convolutional BNNs.
These priors are usually not available in standard BNN frameworks, which is why our novel BNN library, \emph{BNNpriors}, introduces them and makes them easily usable and extendable for research and application purposes.

\section{Description}
\label{sec:description}

The BNNpriors library is written in Python and uses PyTorch \cite{paszke2019pytorch} for defining the neural networks models and performing automatic differentiation.
It offers different inference schemes, including stochastic gradient Langevin dynamics (SGLD) \cite{welling2011bayesian} and Hamiltonian Monte Carlo (HMC) \cite{neal2011mcmc}, but we would recommend using gradient-guided Monte Carlo \cite{garriga2021exact} (which in the library is called \texttt{VerletSGLDReject}).
Moreover, the inference allows for the use of cyclical learning rate schemes \cite{zhang2019cyclical} in order to cover several posterior modes as well as the learning of a preconditioner matrix \cite{wenzel2020good} that scales the momenta of the individual weights.
To the best of our knowledge, this provides the most accurate scalable BNN posterior inference with stochastic gradients available today.

\begin{figure}
    \centering
    \includegraphics[width=0.6\linewidth]{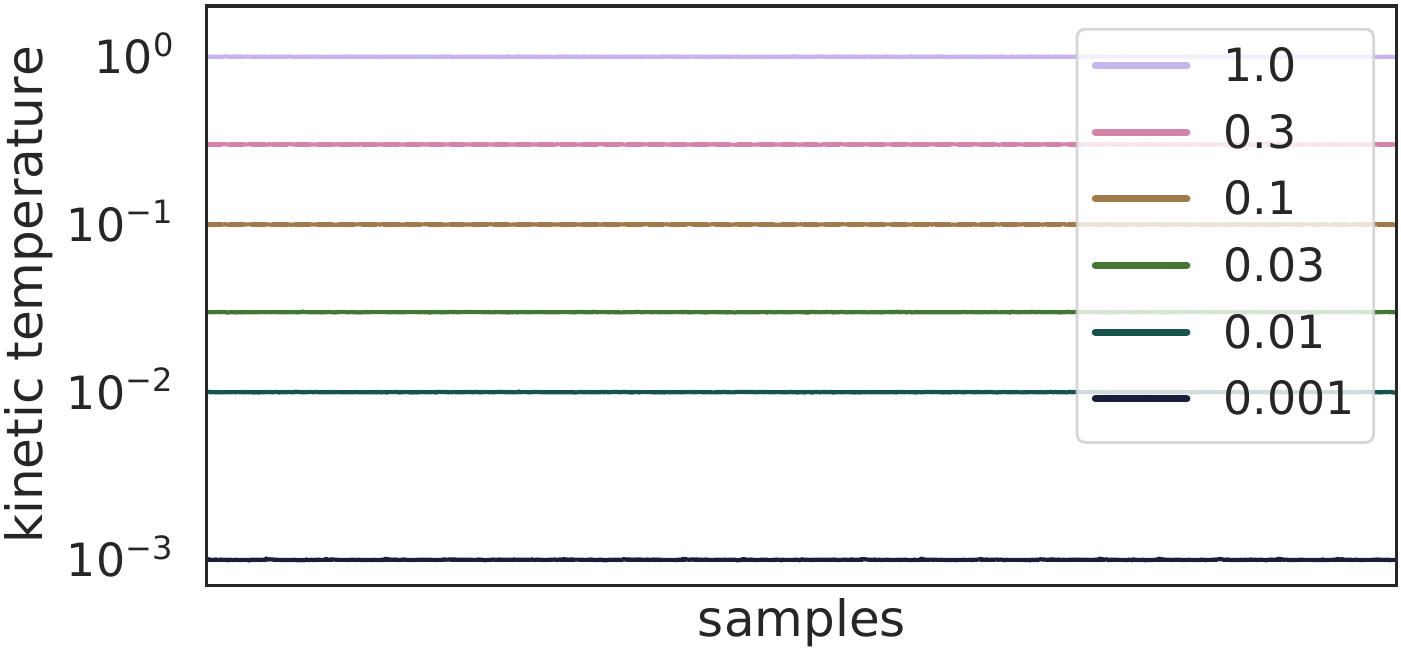}
    \caption{Example plot of kinetic temperature estimates for Markov chains sampled at different temperatures $T$. We see that the kinetic temperatures coincide with the true temperatures and that the inference is thus accurate. (Reproduced with permission from \cite{fortuin2021bayesian}, best viewed in color)}
    \label{fig:diagnostics}
\end{figure}

To perform inference on the standard Bayes posterior, the temperature parameter of the sampler has to be set to $T = 1$. However, one can also use cold posteriors \cite{wenzel2020good}, by setting the temperature to $T < 1$.
In order to assess the accuracy of the inference in every single experiment, the sampler will estimate diagnostics such as the kinetic temperature and the configurational temperature \cite{wenzel2020good}.
These should in expectation coincide with the temperature parameter $T$ and can reveal inference problems, for instance, if the learning rate is set too high or the Markov chain is not converged.
Some example temperature diagnostics of an accurate inference run are shown in Figure~\ref{fig:diagnostics}.

The BNNs in our framework are built from normal PyTorch modules (\texttt{torch.nn.module}), with the difference that their weights are not instances of the \texttt{torch.Parameter} class, but of our \texttt{bnn\_priors.prior.Prior} class.
Our library includes a range of predefined priors within a modular taxonomy, such that new priors can be easily defined and can inherit from existing superclasses, such as location-scale distributions or multivariate distributions.
Moreover, the hyperparameters of our priors can themselves be \texttt{Prior} objects, which allows for the definition of hierarchical prior models.
Also, the \texttt{Mixture} class allows to define mixture priors from all the other existing priors.
A short overview over some popular prior distributions included in our library is given in Table~\ref{tab:priors}.

\begin{table}[]
    \centering
    \setlength{\tabcolsep}{15pt}
    \begin{tabular}{lc}
    \toprule
       Prior  & Density $p(\theta)$ \\
       \midrule
       Gaussian  & $Z^{-1} \exp(\frac{- \| \theta -\mu \|_2^2}{2 \sigma^2})$ \\
       Laplace & $Z^{-1} \exp(\frac{- \| \theta -\mu \|_2}{b})$ \\
       Student-t & $Z^{-1} \left( 1 + \frac{\theta^2}{\nu} \right)^{-(\nu + 1)/2}$ \\
       Cauchy & $Z^{-1} \left( 1 + \frac{\theta^2}{\gamma^2} \right)^{-1}$ \\
       Multivariate Gaussian & $Z^{-1} \exp(- \frac{1}{2}(\theta -\mu)^\top \Sigma^{-1}(\theta -\mu) )$\\
       Multivariate t & $Z^{-1} \left( 1 + \frac{1}{\nu} (\theta -\mu)^\top \Sigma^{-1}(\theta -\mu) \right)^{-(\nu + p)/2}$\\
       Hierarchical & $Z^{-1} f(\theta; \psi)\;$ with $\;p(\psi) \in \mathcal{P}$ \\
       Mixture & $\sum_{i=1}^k p_i(\theta)\;$ with $\;p_i(\theta) \in \mathcal{P}$\\
       \bottomrule
    \end{tabular}
    \caption{Selection of some priors that are available in the library. Here, $\theta$ denotes the weights of the BNN, $\mathcal{P}$ denotes the set of all priors and all the other variables denote hyperparameters of the priors.}
    \label{tab:priors}
\end{table}

\subsection{Usage}
\label{sec:usage}

\begin{figure}
    \centering
    \includegraphics[width=\linewidth]{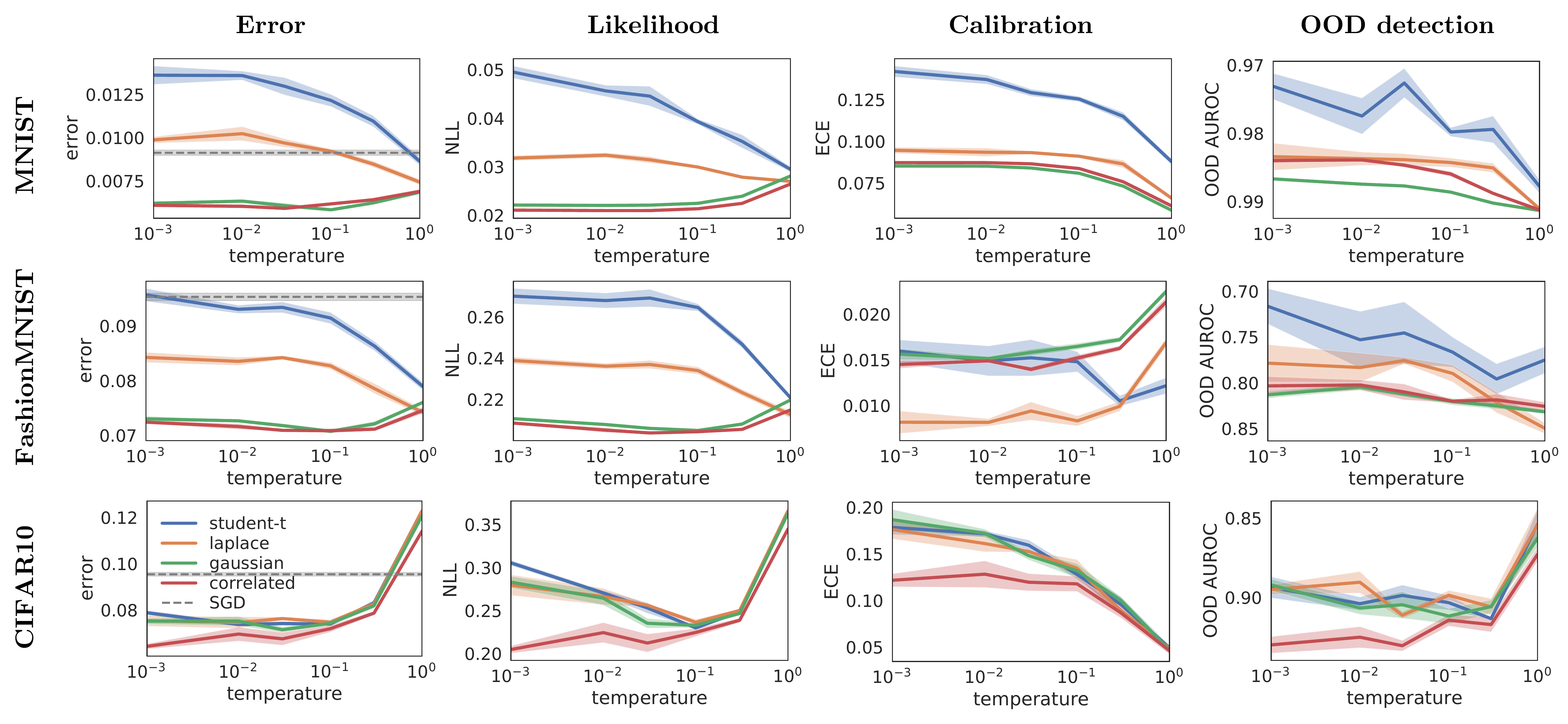}
    \caption{Example tempering curves for different BNN priors on different datasets. We can see that some priors perform much better than others and can also alleviate the cold posterior effect. (Reproduced with permission from \cite{fortuin2021bayesian}, best viewed in color)}
    \label{fig:tempering_curves}
\end{figure}

A basic BNN inference experiment would use the \texttt{train\_bnn.py} script in our library, which takes a number of arguments, including the choice of model, choice of prior, choice of dataset, and some training parameters, such as number of samples, number of training epochs, temperature, learning rate, and similar.
This script will then create an output directory with weight samples from the BNN posterior as well as training curves of different performance metrics and the aforementioned inference diagnostics.
The generated samples can be used with the \texttt{test\_bnn.py} script to create predictions on different evaluation datasets and compute different performance metrics, including accuracy, log likelihood, calibration error, and out-of-distribution detection.
When run with different temperature parameters $T$, one can also plot these performance metrics against the temperature to create tempering curves that can show the cold posterior effect (or its absence).
Some example tempering curves with different priors on different datasets are shown in Figure~\ref{fig:tempering_curves}.

\section{Impact}
\label{sec:impact}

This library has enabled the study of the cold posterior effect in dependence of different priors \cite{fortuin2021bayesian}.
It has led to the first observation that the cold posterior effect in BNNs can indeed be caused by the misspecification of the prior (e.g., by choosing an isotropic Gaussian) and that the performance of the BNN posterior on several metrics can be improved by choosing different priors.
It has helped identify heavy-tailed priors for fully-connected BNNs and correlated priors for convolutional BNNs as better choices.
We expect that these insights will not only spur a series of further studies into the role of priors in BNNs, but also the use of such priors in real-world applications.
Both of these can be catalyzed by our BNNpriors library.

With respect to the inference, our library has been used to show that GGMC inference can use stochastic gradients and still yield nonzero Metropolis-Hastings acceptance probabilities in BNNs \cite{garriga2021exact}, while this is not true for the scheme known as stochastic gradient HMC (SGHMC).
Based on these observations, and in combination with the cyclical learning rates, the preconditioning, and the temperature diagnostics, we thus believe that our library offers the most comprehensive state-of-the-art BNN inference, even without considering the priors.
We hope that his inference can lead to more accurate studies of BNN posteriors and a better performance in real-world applications.

\section{Limitations}
\label{sec:limits}

While our current library is of course limited in the number of predefined priors it contains, we hope that this should not pose a problem in practice, since new priors can easily be defined in our modular framework.
However, a more serious limitation could be that for hierarchical priors, we currently only support joint inference of the hyperparameters with the BNN parameters.
It would be interesting to extend our inference framework to also allow for Gibbs sampling or reversible jump Monte Carlo \cite{green2009reversible} in these models.
Moreover, while our library should generally be usable with any kind of BNN model that is definable in PyTorch, we have not tested it for recurrent neural networks \cite{hochreiter1997long} nor attention-based ones \cite{vaswani2017attention}.
Using our library on such models might require a certain amount of manual tuning.
Finally, in order to aid truly Bayesian model selection of priors, it would be useful to be able to estimate marginal likelihoods from our Markov chains.
This is generally a challenging problem, but there are a few promising solutions \cite{llorente2020marginal, immer2021scalable}.
With these estimates, it could even be possible to learn useful BNN priors entirely from scratch \cite{rothfuss2020pacoh}.
We hope to add all these features to our library in the future, but also welcome open source contributions from the community.

\section*{Competing interests}

The authors declare that they have no known competing financial interests or personal relationships that could have appeared to influence the work reported in this paper.

\section*{Acknowledgements}

VF was supported by a PhD fellowship from the Swiss Data Science Center and by the grant \#2017-110 of the Strategic Focus Area ``Personalized Health and Related Technologies (PHRT)'' of the ETH Domain. AGA was supported by a UK Engineering and Physical Sciences Research Council studentship [1950008].

\bibliographystyle{elsarticle-num} 
\bibliography{references}

\newpage
\section*{Required Metadata}
\begin{table}[!h]
\begin{tabular}{|l|p{6.5cm}|p{6.5cm}|}
\hline
\textbf{Nr.} & \textbf{Code metadata description} & \textbf{Please fill in this column} \\
\hline
C1 & Current code version & \texttt{1.0} \\
\hline
C2 & Permanent link to code/repository used for this code version & \url{https://github.com/ratschlab/bnn_priors/} \\
\hline
C3  & Permanent link to Reproducible Capsule & \url{https://codeocean.com/capsule/8729796/tree/v1} \\
\hline
C4 & Legal Code License   & MIT License \\
\hline
C5 & Code versioning system used & git \\
\hline
C6 & Software code languages, tools, and services used & Python \\
\hline
C7 & Compilation requirements, operating environments \& dependencies & Python 3.7--3.8 with  \texttt{torchvision}, \texttt{torch}~\citep{paszke2019pytorch}, \texttt{sacred}~\citep{sacred}, \texttt{jug}~\citep{jug}, \texttt{h5py}, \texttt{numpy}~\citep{numpy}, \texttt{scipy}, \texttt{tqdm}, \texttt{gpytorch}~\citep{gpytorch}, \texttt{pandas}, \texttt{scikit-learn} and \texttt{matplotlib} \\
\hline
C8 & If available Link to developer documentation/manual & \url{https://github.com/ratschlab/bnn_priors/blob/main/README.md} \\
\hline
C9 & Support email for questions & \texttt{fortuin@inf.ethz.ch} \\
\hline
\end{tabular}
\caption{Code metadata}
\label{} 
\end{table}

\end{document}